\documentclass[lettersize,journal]{IEEEtran}
\def\BibTeX{{\rm B\kern-.05em{\sc i\kern-.025em b}\kern-.08em T\kern-.1667em\lower.7ex\hbox{E}\kern-.125emX}}

\usepackage{times}
\usepackage{gensymb}
\usepackage{tabularx}
\usepackage{lipsum}
\usepackage{array}
\usepackage{booktabs}
\usepackage{listings}
\usepackage{amsmath}
 
\usepackage{enumerate}

\DeclareMathOperator*{\argmin}{arg\,min} 
\usepackage{multirow}

\ifCLASSINFOpdf
 \usepackage[pdftex]{graphicx}
 \else
 \usepackage[dvips]{graphicx}

 \usepackage{amssymb,amsmath}
 
\fi

\usepackage[caption]{subfig}
\usepackage{cite}

\lstdefinestyle{lststyle}{
 captionpos=b, 
 tabsize=2,
 basicstyle=\linespread{0.9}\footnotesize\ttfamily,
}
 
\begin{document}

\title{Federated Learning and Blockchain-enabled Fog-IoT Platform for Wearables in Predictive Healthcare}

\author{Marc~Jayson~Baucas,~\IEEEmembership{Student Member,~IEEE}, Petros~Spachos,~\IEEEmembership{Senior Member,~IEEE}, and Konstantinos~N.~Plataniotis,~\IEEEmembership{Fellow,~IEEE} 
                 
\thanks{This work was supported in part by the Natural Sciences and Engineering Research Council (NSERC) of Canada.
\par M. Baucas and P. Spachos are with the School of Engineering, University of Guelph, Guelph, ON, N1G2W1, Canada.
(e-mail: baucas@uoguelph.ca; petros@uoguelph.ca).
\par K. Plataniotis is with the Department of Electrical and Computer Engineering, University of Toronto, Toronto, ON, M5S3G4, Canada. (e-mail: kostas@ece.utoronto.ca)}}

\maketitle
\begin{abstract}
Over the years, the popularity and usage of wearable Internet of Things (IoT) devices in several healthcare services are increased. Among the services that benefit from the usage of such devices is predictive analysis, which can improve early diagnosis in e-health. However, due to the limitations of wearable IoT devices, challenges in data privacy, service integrity, and network structure adaptability arose. To address these concerns, we propose a platform using federated learning and private blockchain technology within a fog-IoT network. These technologies have privacy-preserving features securing data within the network. We utilized the fog-IoT network's distributive structure to create an adaptive network for wearable IoT devices. We designed a testbed to examine the proposed platform's ability to preserve the integrity of a classifier. According to experimental results, the introduced implementation can effectively preserve a patient's privacy and a predictive service's integrity. We further investigated the contributions of other technologies to the security and adaptability of the IoT network. Overall, we proved the feasibility of our platform in addressing significant security and privacy challenges of wearable IoT devices in predictive healthcare through analysis, simulation, and experimentation.       
\end{abstract}


\IEEEpeerreviewmaketitle

\section{Introduction}\IEEEPARstart{T}{he} usage of wearable Internet of Things (IoT) devices in healthcare is rising. Due to their availability and sensing capability, these devices collect physiological data from patients and provide real-time diagnosis~\cite{wearable-healthcare}. Wearable IoT devices have caused remote healthcare to make a paradigm shift into predictive diagnosis and reliable early detection. The data collected by these devices partnered with different learning techniques aid in predictive healthcare services. Doctors can analyze data such as their patient's activities and accurately predict anomalies and threats against their health~\cite{predictive-health}. They can also prescribe treatments for preventing and addressing these detected concerns. However, this breakthrough has limitations in its technology. Challenge within a network that employs wearable IoT devices cause impasses in predictive healthcare.

An open issue for wearable IoT devices in predictive healthcare is the amount of data it needs to be effective. A large amount of personal data is collected, resulting in security and privacy concerns due to the nature of the data used for analysis~\cite{healthcare-iot-security}. The wearable devices' limitations on processing capabilities lead to vulnerabilities and potential leakages in sensitive patient information~\cite{sp1}. Another issue is the integrity and reliability of the service\cite{healthcare-iot-integrity}. Structuring healthcare to prioritize certain aspects can cause trade-offs in others. Service integrity is crucial for this field in remote healthcare that relies on wearable data accuracy and predictive model precision. One more issue is the adaptability of the network that deploys and serves these predictive healthcare services~\cite{healthcare-iot-adaptive}. Wearable IoT device standardization is a significant concern to IoT networks due to the heterogeneity it introduces. This diversity results in demands for continuous maintenance and updates to the medical server to ensure that it is up to date with every newly introduced wearable IoT device. As a result, concerns about adaptability limit the healthcare network from fully remaining relevant and sustainable over long stretches of its service. 

In this work, we propose a fog-IoT platform to address these issues. We use federated learning to preserve patient data privacy and the integrity of the network's predictive services~\cite{federated-learning-iot}. Also, we incorporate blockchain technology to address the security issues in wearable IoT devices through its access control and cryptographic structure~\cite{baucas1, blockchain-evo}. Finally, we combine these technologies within a fog-based IoT architecture to enforce decentralized servers and resource reallocation to improve the adaptability and sustainability of the overall network~\cite{cloud-fog}. The main contributions of this work are:
\begin{itemize}
    \item We present a fog-based IoT platform using federated learning and blockchain technology to preserve patient data privacy and improve the security of data within the network.
    \item We designed a testbed that simulates and evaluates the proposed implementation. We used model accuracy to observe the platform's ability to preserve the integrity of a predictive service.
\end{itemize}

The rest of this paper is organized as follows. A discussion on the background of our study and a brief literature review are in Section~\ref{back-inf}. Our proposed design and the methodology followed are in Section~\ref{exp}. The presentation of a developed testbed and a discussion of the results are in Section~\ref{eval}. Finally, our conclusions are in Section~\ref{con}.

\section{Background and Related Works} \label{back-inf}
We provide a brief discussion on the relevant works in blockchain technology and federated learning implementations for wearable IoT devices in healthcare. 

\subsection{Benefits of Using Wearable IoT Devices in Healthcare}
Wearable IoT devices can form a network of sensing devices that collect data from points of interest for predictive analysis. In predictive healthcare, these devices collect physiological information from patients for better clinical decisions and to provide a prediction. Each obtained statistic reflects the status of their health, which can identify future issues and potential risks. A benefit of using wearable IoT devices in healthcare is to reduce hospital congestion and medical examination costs via remote diagnoses and long-distance patient monitoring. In~\cite{portable-ecg}, they introduce a wearable Tele-ECG system for heart rate monitoring. The proposed design merges the latest technologies in textile electrodes (TE), Bluetooth low energy (BLE), and smartphones to create a portable means of monitoring and evaluating the condition of a patient's heart for potential anomalies. With features such as geographical tracking, medical history storage, and remote patient monitoring, the system establishes a light and cost-effective alternative for patients suffering from heart issues that need constant observation and evaluation. 

Another advantage of using wearable IoT devices in predictive healthcare is improving the quality of early disease and fault detection in medical centres. In~\cite{biomedical-iot}, they present a review of various biomedical IoT devices that raise the quality of diagnosis in healthcare services. The survey includes wearable sensors as one of the leading technologies that enable advancements in predictive healthcare IoT. Services are now made remote and analytic. Patients can use technologies and wearable sensors that can help give doctors early information on potential causes and triggers of health complications. As a result, there is improvement in predictive anomaly detection without constraints in geographical location and data resource availability. Also, these devices widen the scope of medical centres toward their patients. Doctors can use trend and behaviour analysis to pinpoint anomalies in a patient's health. At the same time, medical professionals can evaluate diseases with improved precision due to leveraged real-time and early detection~\cite{early-detection}. 

\subsection{Challenges of Using Wearable IoT Devices in Healthcare}

Although the usage of wearable IoT devices shows potential to raise the quality of services in healthcare, they also pose the following concerns. 

\subsubsection{Network Security and Data Privacy}
Introducing wearable devices to collect patient data adds more endpoints to the server of the healthcare service. Data collection and monitoring are convenient due to these remote services. However, introducing more devices introduces more vulnerabilities to the network. Increasing the endpoints increases the potential areas where malicious users can attack and steal data. As a result, there is a greater demand for better security as the network grows. Also, privacy becomes a concern due to the sensitive information transmitted from the wearable device to the server. Therefore,  network security and data privacy concerns grow as the network expands with more wearable devices.

Different strategies and technologies have been proposed to address this issue~\cite{forward-data-privacy, deep-learning-privacy}. In~\cite{forward-data-privacy}, they present a multi-keyword search mechanism to preserve data privacy within the IoT network. They aim to strengthen the encryption of the information transmitted from the patient to the medical centre. This strategy achieves forward privacy preservation, which results in a more guaranteed security system for users. Another example that aims for privacy preservation is in~\cite{deep-learning-privacy}. They use deep learning to secure the network by separating private from raw data. This strategy results in minimizing the chances of leaking sensitive information. 

Our approach takes key strengths from these two strategies. We take the ability of cryptographic techniques to reinforce data transmissions and combine them with the adaptability of machine learning techniques to create a secure means of preserving patient data. As a result, we chose blockchain technology for its strong encryption and federated learning for its adaptive ability to improve data privacy while keeping its trained models optimized.  

\subsubsection{Data Integrity and Precision}
An advantage of using wearable IoT devices in healthcare is the real-time diagnosis and early detection of illness and medical anomalies within a patient. However, this can be affected by the quality of the sensor and the data processing scheme behind the service. Potential misdiagnosis is high if the sensor or the evaluation method is not up to standard. Another source of error could come from how the data is collected and stored on the server. Preserving physiological information integrity obtained from the wearable devices is needed as it travels from patient to server. It ensures that the data is recent and precise to the current status of the source. Also, the data handling during aggregation should be carried out with great precision so that the evaluation of the patient is accurate. A service that can not collect, transmit, and process the data correctly and efficiently will result in inaccurate results. Therefore, the integrity and precision of information a concern that needs addressing for a healthcare service that uses wearable IoT devices to be effective. 

Approaches that improve the integrity of sensed data from IoT devices in healthcare services are proposed  in~\cite{data-agg,blockchain-enhance-integrity}. In~\cite{data-agg}, a novel data aggregator for efficient and secure analysis in IoT monitoring is presented. Its scheme involves a cryptographic accumulator that securely collects data from wearable IoT devices. Another example is in~\cite{blockchain-enhance-integrity}. This work highlights the advantages of integrating blockchains with IoT. They aim to enhance the performance of healthcare services that use wearable IoT devices by using the security advantages of blockchain technology. Similar to the approach in~\cite{data-agg}, both strategies focus on the ability of cryptographic technologies to ensure the accuracy of sensed data within the IoT network. 

Our proposed approach differs as it incorporates federated learning for a more secure data aggregator. We combine it with decentralized organizations provided by blockchain and create a platform that ensures a secure medium for accurate data analysis in healthcare. 

\subsubsection{Network Structure Adaptability and Flexibility}
Wearable IoT devices introduce different sensors and technologies that collect and process data. Due to their multiple advantages, healthcare services should include them in their systems. However, innovations and improvements are iterative as changes arise often. Although this is a sign of technological growth and advancements in the healthcare sector, it dictates the pace at which current infrastructures need to keep up. Device diversity has always been a concern for IoT networks due to the need for ongoing standardization. As this new technology is still in development, services that use it should be able to follow the development process. It is the same for wearable IoT devices used in healthcare. The demand for technologies that detect each one is high as new diseases and virus strains often appear. However, every new technology introduced a need for the service to adapt. Networks that are less flexible result in losing their resources. Some even end up losing their service and infrastructure altogether. Therefore, there is a need for adaptability and flexibility in the structure of healthcare networks for the continuously evolving nature of wearable IoT devices.  

Some approaches focus on improving the flexibility of wearable devices to tackle this issue. In~\cite{env-sensing-wrist}, they present an innovative wrist-worn prototype for patient monitoring, which also functions as an IoT gateway. Some sensors in remote healthcare focus on sensing the environment around the patient to ensure that the condition of their surrounding is beneficial to their health. Through it, ambient sensing devices can connect to the healthcare network. As a result, it sets up a platform that can easily integrate and synchronize other devices under it. In~\cite{ecg-patch}, they present a wearable patch that functions similarly to the prototype in~\cite{env-sensing-wrist} by creating a flexible IoT gateway for connecting wearable devices. Instead of ambient sensors, their portal focuses on ECGs and PPGs. 

Our approach is different as we aim to standardize devices through the fog by moving analysis closer to the edge. As a result, we focus on data management standardization instead of device standardization due to the diversity of wearable devices. Instead of specialization through prototypes, we plan to use fog-IoT paired with decentralized technologies such as blockchain and federated learning to develop a modular IoT network for wearable devices in healthcare.

\subsection{Federated Learning for Wearable IoT Devices}
Federated learning is a machine learning technique that takes a distributed approach in training its models~\cite{federated-learning}. It uses its decentralized strategy to utilize global knowledge collected from its clients. In IoT, federated learning has two components; the client and the server~\cite{fl-structure}. Its flow of operations for federated learning starts with each client training their model using their raw data. Then, the server aggregates and compiles the resulting models into a global model that it redistributes to each client's use. As a result, each client receives the most optimal model given global knowledge gathered through local training. This procedure is ongoing, and the global model is updated every time the client provides new knowledge.

Federated learning is well-known for its capability of effectively preserving the privacy of data~\cite{privacy-preserving-fl}. Requiring clients to transmit raw data directly to the server before it is analyzed can cause vulnerabilities. Without a strongly reinforced network, malicious attackers aiming to steal or tamper can target the data. Federated learning protects raw data by creating a strategy that moves the analysis locally. There are lesser avenues for privacy leakages since the server only cares about the resulting trained model from each client~\cite{trustworthy-fl}. As a result, this structure has an improved trusted architecture compared to standard network arrangements. Aside from its security advantages, federated learning also improves the sustainability of IoT networks. It provides a dynamic learning strategy that keeps global knowledge for services up to date~\cite{dynamic-fl}. Commonly, the server aggregates all the data first before it uses it to train the global model. Federated learning allows new correlations and behaviour changes from collected data to be detected by the server earlier. A fully centralized scheme is slowed down and saturated due to the volume of the data it uses to train its models. Federated learning can reduce training time by running it locally and in parallel. Also, there is a significant reduction in the data size as each client trains its model. 

Healthcare services that involve wearable IoT devices revolve around continuous data sensing, learning, and analysis. Federated learning can provide a sustainable and decentralized scheme to optimize these services in healthcare~\cite{sustainable-fl}. We chose to use this technology in our platform due to all the improvements it can contribute to IoT services, specifically healthcare. Our proposed design is different because we aim to reinforce federated learning with the security provided by blockchains. Also, the foundation of our network will be a fog IoT network. The diversity of wearable IoT devices and their limited processing capabilities is a constraint. However, we can overcome this by moving the training to the fog. This arrangement still allows training locally while considering the processing limits of wearable IoT devices. The result is a decentralized IoT network that can maximize the potential of federated learning in keeping data analysis in healthcare services secure and sustainable. 

\subsection{Blockchain for Wearable IoT Devices}
A blockchain is a cryptographic ledger that provides a distributed service for information storage and security~\cite{blockchain-as-ledger}. This decentralized control establishes a trusted system that will only grant access to users acknowledged by the blockchain ledger~\cite{block-survey-ledger}. This feature protects the data from external or unwanted modifications and makes it immutable. For wearable IoT devices in healthcare, the network can use blockchains as a tamper-proof database for patient data storage.

Its unique structure addresses different security issues in IoT networks. In~\cite{block-consent-mech}, they use the unique consent mechanism of blockchains to secure user data privacy in wearable fitness devices. They aim to prove the feasibility of the trustworthiness of blockchains when fortifying the data privacy of wearable IoT device data. Another proposed use for the access control of blockchains is in~\cite{block-auth-comm}. They developed a lightweight authentication scheme that classifies mobile devices and differentiates them from fake data injections or illegitimate users. Their designs show the potential of blockchain technology in securing the healthcare server and the monitoring devices used by its services. In~\cite{block-encrypt-search}, they use a different feature of blockchains in reinforcing their healthcare service that uses wearable IoT devices. They integrate the encryption model of blockchains to improve the security of the IoT network. Their design creates a searchable encryption technique that assists in securing collected COVID-19 data. It shows the effectiveness of blockchain architecture in helping protect the healthcare IoT network from several security threats like malicious data injection and hijacking sessions. 

Our design differs from the previously presented implementations because we aim to use a private blockchain~\cite{baucas2}. Public blockchains use a reward-based protocol called ``proof-of-work" (PoW) in granting access to their devices~\cite{block-survey}. Although this protocol is secure, its processing requirement is a caveat for wearable IoT devices in healthcare. In~\cite{baucas-mag}, we highlighted how some wearable IoT devices are incapable of adapting to this authentication structure due to their low-cost and low-end designs. Therefore, we need a different blockchain architecture that can cater to the design nature of wearable devices. Private blockchains function differently. They incorporate a more trust-based protocol. This approach turns the blockchain into a hyperledger that can regulate its devices based on a list of trusted devices defined by the network owner. Healthcare IoT networks can benefit from this approach more since the need for high processing power for their wearable devices is reduced. As a result, the types of IoT-based wearable devices they can incorporate into their services do not limit the healthcare network~\cite{biothr}.

\section{Design and Methodology} \label{exp}
The following is a discussion of our proposed platform, including a description of the design and the different components.

\subsection{Architecture Overview}
We propose a fog-IoT-based approach to secure data exchange from wearable IoT devices used for predictive healthcare services. We aim to improve the overall structure's service integrity and flexibility. To further enhance the distributive organization of our network, we chose to use federated learning. With its ability to aggregate learning models, we can sort our data and keep private information secure. Also, it can reduce the sources of leakage since each client is only required to send their locally trained model to the server. Integrating federated learning improves the sustainability of the predictive healthcare service by providing a system that can organize the model data and provide a global model that represents the overall knowledge obtained through all the local training. The result is a scalable design that introduces flexibility by keeping the network distributed while ensuring that it does not impede the learning and analysis of the service. 

Next, we combine federated learning with private blockchain technology for more secure client authorization. Also, it is treated as a hyperledger to store the IDs of each client device and locally trained model. Due to its immutability, tracking and logging each trained model for version control and debugging improves the data analysis and learning aspect of healthcare networks and their services.

Commercial wearable devices are diverse. Some are low-end by design. To address this, we moved the process of locally training the model from the edge of the network to the fog. We can reallocate the training process by providing an intermediary fog device. Usually, training takes a large portion of processor capacity. Unfortunately, not every wearable IoT device can do it effectively. For our platform to be able and handle a diverse pool of wearable devices, we offloaded the processing required to a fog device. 

\begin{figure}[t!]
\centering
\includegraphics[width=\columnwidth]{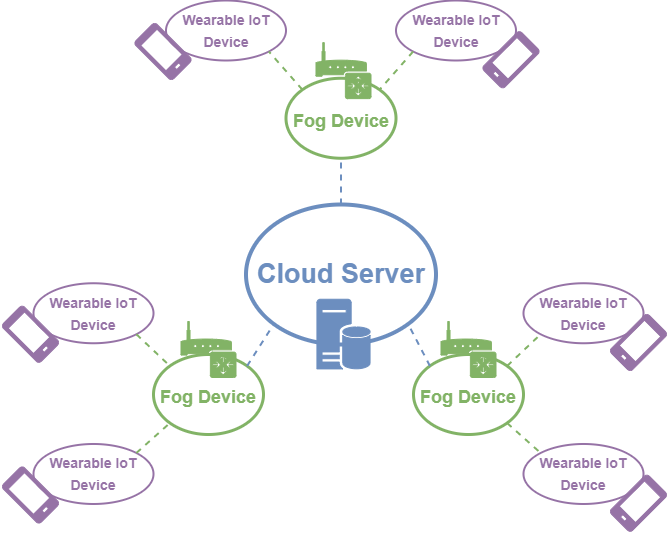}
\caption{Network arrangement for the proposed fog-based platform for wearable IoT devices.\label{netset}}
\end{figure}

The proposed network arrangement is shown in Fig.~\ref{netset}. First, the fog device collects all sensed data from each wearable IoT device and trains the model locally. Each one sends its local model to the server for storage and tracking. Next, the server aggregates all the knowledge from the local models and generates a global model. Then, the server sends this model to each wearable IoT device through the fog nodes for further data analysis. We investigated the feasibility of our platform by simulating the intended data flow of our design and the effectiveness of our federated learning system. Lastly, our implementation aims to provide a low-cost setup that simulates our platform under a resource-constrained environment. We presented a top-down view of our architecture and its data flow. The following details will encompass the other design choices we made for further establishing the predictive healthcare service that our design models.

\subsection{Dataset and Neural Network}
To establish the predictive healthcare service our platform will use to test its feasibility, we looked for a dataset and neural network configuration that fits our remote monitoring data flow. We chose a standard classifier design and organized the dataset to minimize its impact on our platform experiments. Our focus is on the effectiveness of our platform in securing wearable IoT device data and ensuring the integrity and flexibility of the healthcare service, and not maximizing the accuracy of the classifier. Therefore, the classifier we used in testing our platform is for human activity recognition (HAR). It uses accelerometer data from various commercial wearable IoT devices such as smartphones and smartwatches to determine the posture and condition of each user~\cite{har-in-health}. Healthcare services use these to monitor the safety of their patients. For example, it can detect the falling pattern of a person and enable real-time response. Also, they can monitor people's daily exercise and regimen for further consultation and improvements for their health~\cite{har-use-exercise}. Similarly, we will use it to simulate patient data for training HAR models to better reactive analysis and accident detection in wearable IoT device-based remote monitoring. 

The dataset we selected is the Human Activity Recognition Using Smartphones Data Set from the UCI Machine Learning Repository\cite{uci-har}. It is from an experiment with 30 volunteers within 19-48 years. Each individual performed six activities: walking, walking upstairs, walking downstairs, sitting, standing, and lying down. Each volunteer did each action wearing a Samsung Galaxy SII smartphone on their waist. The result is a dataset that contains 10299 instances of 3-axial linear acceleration and 3-axial angular velocity captured from the accelerometer and gyroscope of the mobile device. It encompasses 561 labelled features with time and frequency domain variables. We chose this classifier and dataset combination because the data is from smartphones, which is more common to the public. This selection provides a better dataset scope that represents a larger population. Our focus is on the federated learning aspect of the design. We can minimize any impacts of extraneous variables that could interfere with our experiments and the precision of its results by using an already tested and documented classifier. Using it provides convenience and efficiency as we simulate the HAR-based predictive service testbed for our experiments. 

We implement a one-dimension convolutional neural network (CNN) because the dataset is a sequence with only a single-dimensional source. Also, it is a design that minimizes the overall processing requirement when training and testing the model. Since we aim for a low-cost approach, we intend to limit the impact of resource requirements. We coded it using a combination of the Keras and Tensorflow libraries in Python. The dataset was already pre-grouped to have a 70-30 split for training and testing. We use the same split when training and testing our CNN to avoid generating any significant impact from the nature of the classifier or the data. Also, we refrained from potentially over-tuning the CNN by only calibrating through the epoch number. We wanted to ensure that the experiments focused on the performance of the federated learning aspect of our design.

\subsection{Federated Learning Implementation}
We implemented the federated learning aspect to incorporate distributive learning with the HAR predictive model. We used the previously presented CNN as the learning model for our cloud and fog device. The target of the adopted CNN is to learn from various human activities using accelerometer and gyroscope data. The result is a HAR model for potential anomaly detection in patient movement and activeness. Federated learning aims to solve for an optimal global model $w_{G}$ that minimizes the weighted average loss of all clients. It takes the global loss function $\mathcal{L}$ and simplifies it into a summation of losses obtained from the local models $w_{L}$ of each client shown in the following:

\begin{equation}\label{fl}
  w_{G} = \argmin_{w_{L}} \mathcal{L}(w_{L}) = \argmin_{w_{L}} \left[\frac{1}{K} \sum_{k=1}^{K} \mathcal{L}_{k}(w_{L})\right],
\end{equation}

where $\mathcal{L}_{i}$ is the local loss function by client $k$. We translated the solution for this federated learning problem as a python script executed by the central server. It starts by iterating through each local model we obtained through the fog devices.

\begin{figure}[t!]
\centering
\includegraphics[width=\columnwidth]{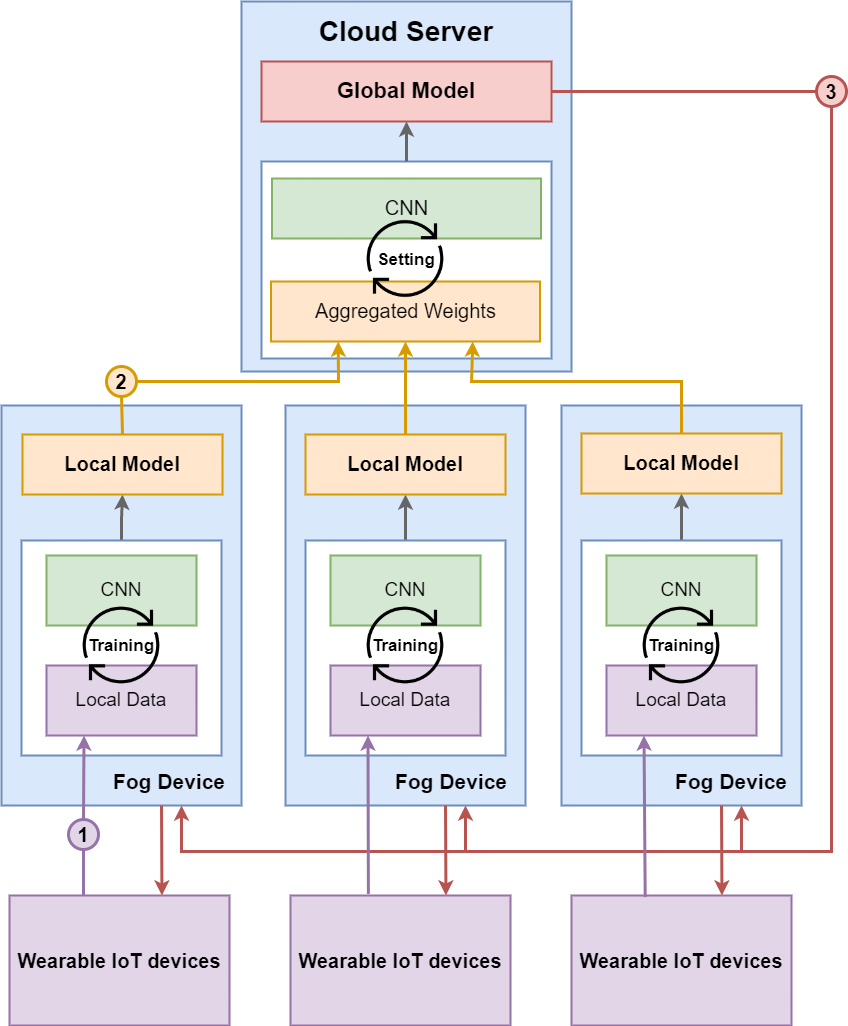}
\caption{Flow of operations for our implemented federated learning system within our platform.\label{fedlearn}}
\end{figure}

Next, we extract the weights of each model and optimize them by scaling them according to the total number of clients and their accuracy. We optimized our weights by giving the model with the highest accuracy the best scaling factor and gave the rest lower scale values relative to the number of total fog clients in the network. This scaling scheme ensures that the best-performing local model is the base, and the rest are additional references for further learning reinforcement. While scaling, the server generates a CNN with the same layer design. Then, the script aggregates the optimized weights via averaging, takes the resulting solution, and sets it into the prepared model, which becomes the global model. This model encompasses all the knowledge obtained from each locally trained CNN. A graphical representation that shows the flow of our federated learning system is in Fig.~\ref{fedlearn}. 

\subsection{Private Blockchain Implementation}
We incorporated a private blockchain to manage the training information shared by the devices that will train the HAR-based predictive model. The blockchain protocols create an access level ensuring that only trusted devices can access the predictive model's local training and global knowledge. We coded it in Python. Each block is a class definition that contains an ID, a list of training model results, and a hashed string of the previous block. The private blockchain authorizes communication between the server and the client based on their assigned ID. Also, it collects the locally trained models and stores them periodically within a block. These blocks are chained cryptographically by generating a hash of the previous block and storing it within the new one. The resulting data structure is a log of every training model sent to the server and the other actions the network carried out. The blockchain becomes a hyperledger that stores the history of every operation done by the network. Also, since it is tamper-proof, these records are protected. We also initialize copies of the blockchain and distribute them to each client to keep the data decentralized and free from manipulation outside the authorized devices. This distribution allows the service to check the validity of each client and the local models they are training and sending to the server.

With the private blockchain, the information and operation within the network remain secure. Tampering becomes more difficult with the hyperledger keeping track of all the changes to the global model. It can record the results of each local training through the obtained models from the fog clients. Also, private and raw data from the clients are not sent to the server, keeping them secure from any form of leakage or malicious attacks. Overall, the blockchain sets up the network by securing its endpoints and reinforcing its decentralization.

\subsection{Hardware and Software Specifications}
To simulate the predictive service based on the HAR model, we selected hardware that fits our aim to have a low-cost design. Our setup makes use of Raspberry Pi 3 Model Bs. We chose Pis due to their portability, programmability, and modularity for rapid programming. Also, we selected a low-cost and low-end device to simulate the capabilities of most wearable IoT devices. Each Pi has a Quad-core ARM Cortex A53 processor that runs on a benchmark of 1.2 GHz. Our focus is on model accuracy while processing speed is currently not a concern. We also used Pis for both the fog devices and the central server of our design. Each Pi is pre-loaded with a Raspbian-Jesse operating system. We installed Python 3.6 on every Pi as our programming language due to its flexibility and diverse selection of open-source labels. The selected version is the most recent version that works with all the Python libraries needed for our design. As for the wearable IoT device that represents the edge of our network, we based it on the same smartphones from the HAR dataset we chose to use. It is a Samsung Galaxy SII that uses a Dual-core 1.5 GHz Scorpion processor.

\section{Results and Discussions} \label{eval}
We tested our proposed platform to evaluate its feasibility by addressing data analysis security, integrity, and flexibility. 

\subsection{Testbed}
We structured our testbed to simulate data flow from the fog to the cloud server. It is to observe the behaviour of our platform under the standard network interactions of local and central servers of a fog-IoT network. In designing our testbed, we use low-cost and low-end devices to evaluate our design under minimal processing capabilities. We chose this approach to show the feasibility of our proposal when using simple IoT devices. It gives an estimate for a benchmark on the minimum system requirements for our implementation. We decided to use Raspberry Pi 3 Bs to model the fog devices because it is modular, portable, and low-cost. During the early stages of implementation, we discovered that the Pi could run the classifier training and the blockchain scripts without any issues. Therefore, we continued to use Pis as both the fog and the cloud servers in our network. Another reason why we chose to stay with just one type of technology was to limit the potential impact of any processing or data capacity advantage if we used a device with a better processor as a server. Since Pi's are modular, we replicated each node by flashing an identical copy of the operating system to each device. As a result, we arranged the network to have a Pi as the cloud server with 10 Pis that serve as the fog nodes. 

First, we assume that each fog server already contains the wearable IoT device data, so the training and testing dataset is pre-loaded in each Pi. Since our focus is on the federated learning aspect carried out by the fog-IoT servers, we can skip the simulations of data transmissions between the wearable and the fog devices. Next, each fog device will train its models locally and send them to the central cloud server via wireless file transfer. The server will wait until it receives all the local models before the next step. Next, it will aggregate the weights of each model and scale it according to its accuracy. We implemented a scaling scheme that gives the highest possible scaling factor to the local model with the highest accuracy. At the same time, the rest of the models will receive the same scaling factor but are significantly lower. Then, the server will take the resulting scaled weights and set them on a structurally identical CNN, resulting in the global model. For our test, we plan to compare the accuracy of the global model with the average of the locally trained models from each fog device. We elected to use this metric to evaluate the behaviour of our platform. Also, we aim to ensure that our implemented federated learning system does not disrupt the integrity of the classifier. Finally, to investigate the security and adaptability of our design, we later discuss the strengths of our platform that we observed as we tested and simulated the platform.

\subsection{Integrity Evaluation via Training Model Accuracy}
Our test investigated the ability of our platform to maintain the integrity of the predictive data analysis of the classifier within the healthcare network. We used a 1D CNN to classify HAR using the HAR dataset from the UCI Machine Learning Repository. We chose to evaluate the feasibility of our design by comparing the accuracy between the global and the local models. 

\begin{figure}[t!]
\centering
\includegraphics[width=\columnwidth]{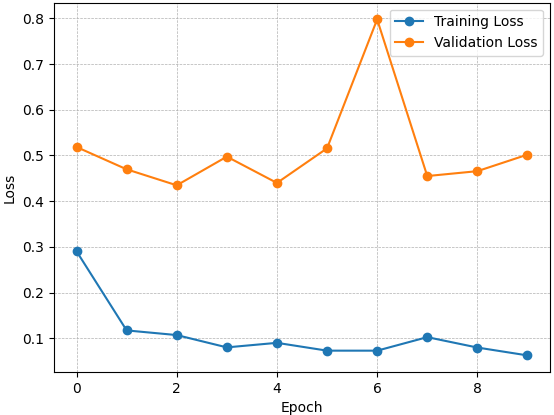}
\caption{Sample training and validation loss of one local model yielding a testing accuracy of 90.43\%.\label{loss}}
\end{figure}

\begin{figure}[t!]
\centering
\includegraphics[width=\columnwidth]{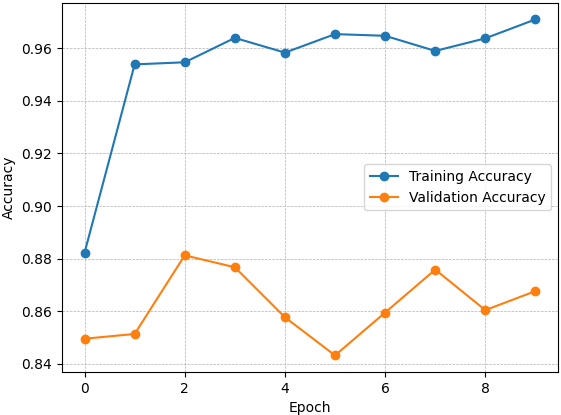}
\caption{Sample training and validation accuracy of one local model yielding a testing accuracy of 90.43\%.\label{acc}}
\end{figure}

First, we trained the model locally using 10 Pis pre-loaded with the classifier and the dataset. We arranged the models by assigning a number to each corresponding client. The fog devices trained each local model using the same 70\% split of the dataset. Also, we tested each model with the remaining 30\% of the dataset. We tried to minimize the complexity of our classifier and focused on the performance of our federated learning system by doing the minimal tuning. The only parameters we tinkered with were the epoch number and the batch size, which we set to 10 and 8, respectively. The resulting model yielded a 90.43\% testing accuracy. We generated its validation-to-training loss and accuracy plots as shown in Fig.~\ref{loss} and Fig.~\ref{acc}, respectively. These results are a sample of 1 out of 10 fog clients. Each presents the reasonable validity of the classifier at minimal tuning in correctly identifying the different labelled human actions based on accelerometer and gyroscope data. 

Upon further analyzing these graphs, we observed a spike in the validation loss on epoch 6 of Fig.~\ref{loss}. This behaviour is due to the learning rate causing the trained model to be volatile at this point. Since the graph stabilizes, it does not significantly impact the overall model. Also, there is a similar spike on epoch 5 of Fig.~\ref{acc}. This behaviour could reflect potential volatility within the dataset. However, since the amplitude of the fluctuations is within a reasonable range of +/- 2\% from the median, it shows that the data's volatility is not significant enough to render the model inaccurate.

\begin{figure}[t!]
\centering
\includegraphics[width=\columnwidth]{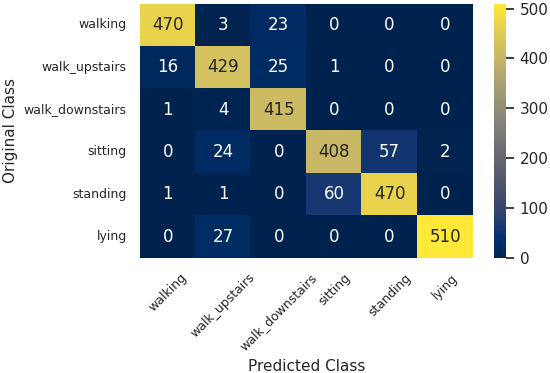}
\caption{Confusion matrix of testing the global model with aggregated weights from 10 local models.\label{confu}}
\end{figure}

Overall, each trained model yielded similar behavioural trends but had varying final accuracy values within a range of +/- 1.28\% from the median. This range shows the impact of the data on the accuracy of the model. However, it is not significant enough to invalidate the training results. Therefore, we took the average of their performances and used it to represent the benchmark of the classifier for each collective number of clients. After getting the average accuracy of the local models, we compared it with the global model to evaluate our implemented federated learning system.

First, we generated the global model for each N number of clients. We have a visualization of our global model's performance after aggregating the weights of local models from 10 clients through the confusion matrix in Fig.~\ref{confu}. Next, we compared the average accuracy of the N local models to their corresponding global models. This method allows us to see the model's ability to keep up with the average performance of a collection of locally trained models.
We obtained the average accuracy of the local models while increasing the number of clients. Then, we generated their corresponding global model and recorded its accuracy.

\begin{figure}[t!]
\centering
\includegraphics[width=\columnwidth]{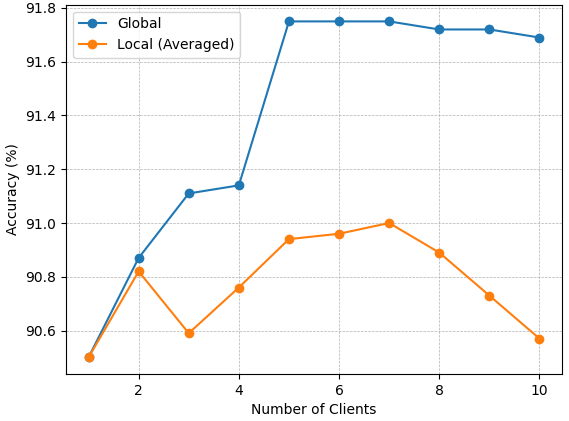}
\caption{Difference in testing accuracy between global model and average local model as the number of clients increases.\label{accmodel}}
\end{figure}

Finally, we plotted the comparison between the accuracy of the global model and the averaged local model shown in Fig.~\ref{accmodel}. Based on the results, we can observe that as the number of clients increases, the accuracy of the global model keeps up and even eventually performs better than the average accuracy of its local models. Through this observation, we can see the ability of our implemented federated learning system to minimize any loss from aggregating weights. We can attribute this improvement in accuracy to our optimization scheme that assigns a higher scaling factor for the local model with the best performance. Also, we can see how the accuracy of the global model peaks at 91.75\%.

We can conclude that our scaling scheme can maximize the classifier's performance by prioritizing the local models that yield the best results. Also, the platform performed effectively even with the low-cost devices functioning as servers. Therefore, we can further reinforce the argument that our design can be scalable and efficient even under resource constraints. Since the Pis are modular and portable, our platform can cover remote monitoring applications constrained by distance and mobility. Overall, this shows the potential of our design as an effective and scalable approach that can benefit wearable IoT device-based predictive healthcare even with resource constraints.     

\subsection{Security and Adaptability Analysis}
The following is an analysis of the security and adaptability of our design. Each evaluation is from observations from our simulations and testing:

\subsubsection{Security}
Our design choices aim to preserve the data privacy of wearable IoT devices. Federated learning allows us to keep raw data from leaving the local network. Since only the models are transmitted, limiting potential leakages protects the data of wearable IoT devices. With blockchains, our platform can create a more exclusive network where only trusted devices can send data to the cloud server. As a result, the hyperledger can reduce the endpoints that expose the server to malicious attacks such as spoofing and impersonation. Also, this can reduce the threat of Denial of Services (DoS) attacks. Since the blockchain can regulate which devices can actively communicate with the server, it can serve as a rate limiter for any targeted attack. Blockchains provide an extra level of security due to their built-in access control and cryptographic structure. Overall, each technology integrated into our platform can preserve the data privacy of wearable IoT devices.


\subsubsection{Adaptability}
Based on our observations, our fog-based IoT architecture was able to aid in implementing both federated learning and blockchain technology with ease. We can attribute it to the decentralized structure of fog-IoT. Due to the fog architecture, both technologies can perform optimally under this network arrangement and utilize their strengths. Device diversity causes network strains due to the constant demand for updates and maintenance. However, federated learning reduces this strain by standardizing transmitted information into a uniform package. This package is the locally trained model. Lessening the demand for the cloud server to keep up with the latest wearable device allows it to focus on deploying services efficiently while keeping up with the most recent global knowledge. As long as the classifier structure remains uniform, the server only needs to regenerate the same model with different weights. As a result, the regeneration and aggregation become modular to the service. This modularity contributes to the adaptability of the network. It enables the use of other classifiers and wearable IoT device groups.

Keeping the network decentralized through the fog also allows better clustering. This design reduces the time required to update all its servers and devices. For instance, take a heterogeneity equation $H$ based on the distribution of worker update time $\phi_w$ presented in the following: 

\begin{equation}\label{het}
  H = 1 - \frac{1}{W-1} \sum_{w=1}^{W-1} \frac{\phi_{W}}{\phi_{w}},
\end{equation}

According to Eq.~\ref{het}, a higher worker count $W$ lowers the overall heterogeneity $H$ of the network. Also, we assume that $\phi_W = Min(\phi_1 ... \phi_w)$. This assumption means that the heterogeneity value of the network is proportional to how spread out is the update times of each worker. The higher this value $H$, the lesser the network is impacted by the updates times of its devices. In a standard cloud IoT network, the number of workers equals the number of wearable IoT devices it holds. As a result, the heterogeneity value will be high due to the diversity of the wearable devices connecting to the server. With varying update times, the $\sum_{w=1}^{W-1} \frac{\phi_{W}}{\phi_{w}}$ term will be asymptotic to 0. When plugged in the rest of the equation, the heterogeneity value is maximized and asymptotic to +1. The closer this value is to 1, the more heterogeneous the network. Our fog-based IoT approach allows us to reduce the impacts of the update time of each worker by standardizing them. Instead of having each wearable IoT device as a separate worker, they are clustered by the fog device and become the new worker. With standardized devices, the $\sum_{w=1}^{W-1} \frac{\phi_{W}}{\phi_{w}}$ term will be asymptotic to $W$. When plugged in the rest of the equation, the heterogeneity value is minimized and asymptotic to -1. The closer this value is to -1, the less heterogeneous the network. Ideally, we aim to minimize the heterogeneity value to indicate a network less affected by the diversity of its worker's update times. 

With our experiments, the update time is each fog device's training and communication time. We can observe in the testbed design that having a distributive arrangement can reduce the impact of the diversity of wearable IoT devices on the overall update time by offloading the training and processing to the fog. Instead of many wearable IoT devices that can yield varying update times, sending training data to the cloud server can be standardized. So theoretically, we can infer that our platform will result in a lower heterogeneity value. This analysis highlights the benefits of the distributive approach that we proposed. 

Also, the resulting accuracy of the global model presents how the precision of the predictive healthcare service is not affected by the shift in the location of the learning process. Since the edge device is only required to send data securely, it does not need standardization as long as the fog device is aware of the data it receives. Also, having one less complex process within a sensing device makes its design more robust. Removing this constraint helps the server keep up with the constant introduction of new wearable IoT devices by simplifying the data flow within the network. This change creates a more adaptable network that can cater to a diverse pool of wearable IoT devices. 

\subsection{Future Work and Recommendations}
Further improvements to the design can be using other metrics for testing. Another performance metric we can include for future iterations can be the propagation delay between the fog and the cloud when transporting the training model results. This addition introduces potential dynamic scenarios that further test the mobility and reach of our platform. Another metric could be the power consumption of the servers. The coverage of servers must reach remote areas in extreme cases of predictive healthcare. As a result, more portable and power-efficient designs are in demand for sustainability and longer service uptime. This design addition introduces the potential design and device optimizations that make the platform more cost-efficient while managing the predictive healthcare service. Another improvement is to test the platform with more than one neural network design and dataset. This addition can further emphasize our platform's modularity and performance under different configurations.
 
\section{Conclusion} \label{con}
We propose a platform that addresses the issues of data privacy, service integrity, and network structure adaptability of wearable IoT devices in predictive healthcare. We used federated learning for its ability to effectively aggregate local models into a global entity to ensure the integrity of the predictive service. We further incorporated private blockchain technology to reinforce the overall network security. Lastly, we have fog-IoT as the base for offloading and process redistribution. By evaluating the implemented federated learning system in terms of model accuracy, we observed its feasibility in maintaining the integrity of the HAR classifier. Next, we discussed the effectiveness of our platform even when using a low-cost and low-end device as its fog and cloud server. Then, we analyzed our design choices and highlighted its strengths in terms of privacy preservation, security, and network adaptability. Overall, through our testing and evaluation, we saw the feasibility and potential of our proposed platform in addressing the security, integrity, and adaptive issues of wearable IoT devices in predictive healthcare.  
\bibliographystyle{IEEE}
\bibliography{IEEEabrv,federated}

\end{document}